\documentclass{article}
\usepackage[T1]{fontenc}
\usepackage{iclr2027_conference,times}
\iclrfinalcopy
\usepackage{microtype}
\usepackage{graphicx}
\usepackage{subcaption}
\usepackage{booktabs}
\usepackage{amsmath}
\usepackage{amssymb}
\usepackage{mathtools}
\usepackage{xcolor}
\definecolor{linkcol}{RGB}{0,45,110}
\usepackage[colorlinks=true,allcolors=linkcol]{hyperref}
\usepackage{url}
\usepackage[capitalize,noabbrev]{cleveref}
\definecolor{relaycol}{HTML}{C99A3C}
\newcommand{\lossbar}[1]{\textcolor{relaycol}{\rule{#1\dimexpr 2.6em\relax}{1.2ex}}}

\renewcommand{\iclrruler}[1]{}

\title{Agents Cannot Hand Over What\\ They Have Not Yet Understood}

\author{Yibo Liu\\
Beijing University of Posts and Telecommunications}

\begin{document}
\maketitle
\lhead{}\renewcommand{\headrulewidth}{0pt}
\raggedbottom

\begin{abstract}
Agents that work beyond a single context window, or in teams, hand their work to a successor as a short written
summary. Evaluation has focused on whether an agent can solve a task, and relatively little is known about how
much of what an agent has learned survives being written down. We study this in 56 interactive worlds
with hidden rules, where an agent that is told nothing fails and an agent given a 300-character statement of the
rules succeeds. We compare one agent that plays eight lives and keeps its whole history with a chain of eight
agents that each play one life and inherit a 300-character note. All six models we test learn less in
the chain: after eight generations it stands where the single agent stood after about four lives.
Controlled variants of the chain decompose this \emph{handoff loss} into three components: none from changing the
agent, a smaller loss from compressing to 300 characters, and a larger loss from compressing repeatedly. When
each note is written from the full record of earlier lives instead of from the previous note, the same 300
characters recover 70\% to 82\% of the loss. The loss is robust across model sizes and is not
alleviated by instructing the writer to record evidence, to revise the inherited note, or to keep a structured
ledger. In simpler terms, an agent cannot hand over what it has not yet understood: a note holds what its writer
has worked out, and a rule that takes several lives to see is lost with the lives that were not passed on.
\end{abstract}

\section{Introduction}
\label{sec:intro}

Language-model agents now work on tasks that outlast a single context window, and on tasks divided among several
agents \citep{yao2022react,wu2023autogen,hong2023metagpt,li2023camel}. Both situations force the same operation:
an agent compresses what it has learned into a short text, and a successor continues from that text in place of
the experience. Context compaction does this within one run
\citep{packer2023memgpt,kang2025acon,li2026compactionrl}, reflection and long-term memory do it across episodes
\citep{shinn2023reflexion,zhao2023expel,wang2024awm,xu2025amem,chhikara2025mem0}, and a sub-agent's report or one
agent's takeover of another's unfinished work does it across agents \citep{kc2026handoff,cemri2025mast}. We call
this operation a \emph{handoff}.

However, relatively little is known about what a handoff loses when the work is unfinished. A failed handoff has
two possible causes that existing evaluations cannot tell apart: the writer never understood the task, or it
understood and did not say so. On real tasks what the writer knew is not observable. Studies of handoff and
compaction therefore measure downstream cost or task completion
\citep{kc2026handoff,wang2026lost,liu2026compressioncost}, and memory benchmarks measure recall of facts the
agent was given \citep{maharana2024locomo,wu2024longmemeval,hu2025memoryagentbench,ai2025memorybench}. Neither
says how much of what an agent worked out for itself survives being written down.

\begin{figure}[t]
  \centering
  \includegraphics[width=\linewidth]{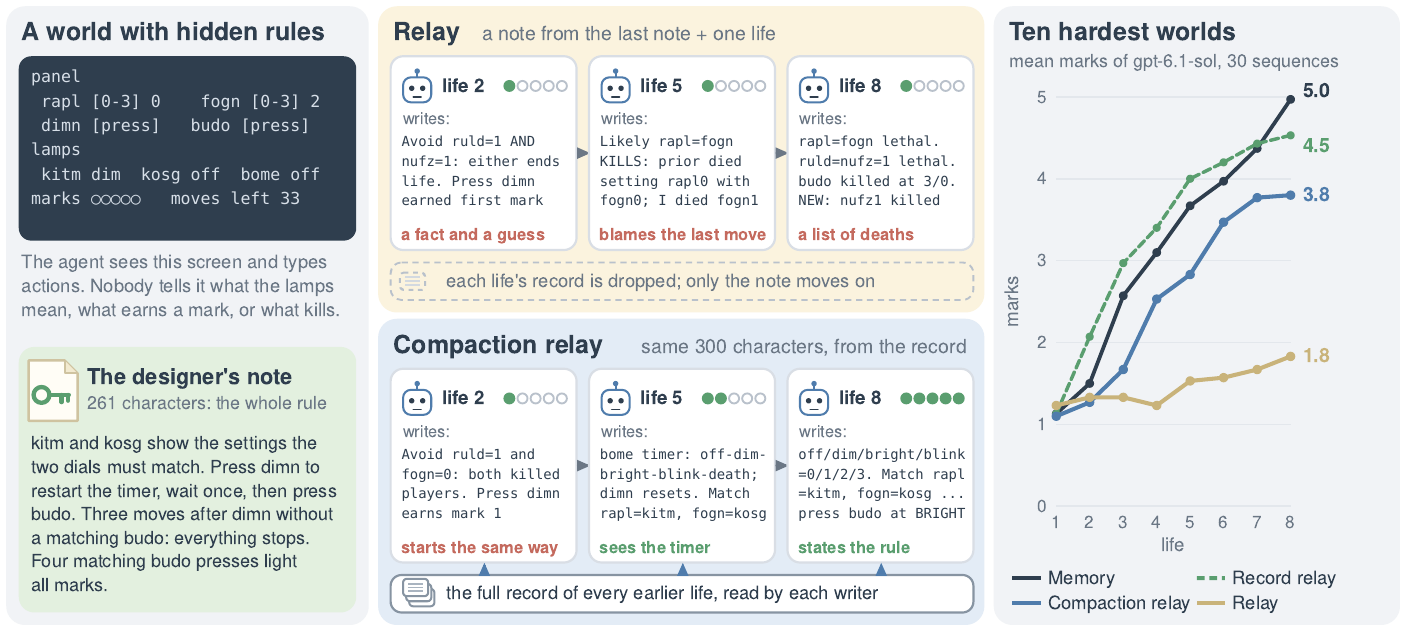}
  \caption{\textbf{An agent cannot hand over what it has not yet understood.} \emph{Left}: one of our worlds as an
  agent sees it, and the designer's note, which states its whole rule in 261 characters. \emph{Middle}: notes
  written by gpt-6.1-sol on this world (excerpts; dots are the marks reached in that life). In Relay each note is
  written from the previous note and one life, and the chain collects guesses about what ended a life. In
  Compaction relay the note has the same limit but its writer may read the record of every earlier life, and the
  rule is found. \emph{Right}: mean marks per life on the ten worlds where the loss is largest.}
  \label{fig:overview}
\end{figure}

We study this in worlds where the knowledge to be handed over has a ground truth (\cref{fig:overview}). Each of
our 56 worlds is a small interactive program with hidden rules. An agent that is told nothing fails
within one life; an agent given a 300-character statement of the rules succeeds almost always; and because the
rules are a finite list of facts, any note can be graded fact by fact. We compare one agent that plays eight
lives and keeps its whole history (\emph{memory}) with a chain of eight agents that each play one life and
inherit only a 300-character note (\emph{relay}). The two meet the same situations in the same order. The design
follows a human experiment in which chains of players passed messages about unfamiliar games and learned about
as fast as individuals with unlimited lives \citep{tessler2021growing}, in the tradition of transmission-chain
studies \citep{bartlett1932remembering,mesoudi2008multiple,kirby2008cumulative,caldwell2008experimental}.

All six models we test learn less in the chain than alone. After eight generations, relay stands where
the memory agent stood after about four lives; between 19\% and 48\% of what a model
gains from its history does not pass through the note (\cref{sec:curve}). We call the difference the
\emph{handoff loss}. It is absent in worlds that a model learns within two lives and in worlds it cannot learn at
all, and largest in between, so each model loses on the worlds at the edge of its own ability
(\cref{sec:where}). A stronger model does not remove the loss. It moves it to harder worlds.

Memory and relay differ in three ways at once: the agent changes, the past is cut to 300 characters, and each
note is written by someone who has seen only the previous note and one life. Two further chains take these apart
(\cref{sec:interventions}). When each new agent receives the complete record of earlier lives, it does as well
as the memory agent: changing the agent costs nothing. When each agent still plays from a 300-character note but
the writer of that note may read the complete record, the chain recovers 70\% to 82\% of the
loss on the worlds where it is largest, and its notes state about twice as many of a world's facts as ordinary
notes do (47\% against 24\% for gpt-6.1-sol). Most of the loss therefore comes from compressing repeatedly, and only the remainder
from compressing at all.

The loss is robust across model sizes and is not alleviated by telling the writer what to do. Asking for
evidence, asking for a revision of the inherited note, and prescribing a ledger of what is confirmed, ruled out
and open change little. A longer note does help, because it carries raw observations forward; a stronger writer
with the full record needs no extra room. In simpler terms, an agent cannot hand over what it has not yet
understood. A note holds what its writer has worked out. A rule that takes several lives to see is lost with the
lives that were not passed on, a version in language-model agents of the loss of evidence that iterated learning
predicts for chains of learners \citep{griffiths2007language,beppu2009iterated}. The practical consequence is
that a summary of unfinished work should be rewritten from the record each time, not built on the last summary.

Our contributions are as follows.
\begin{itemize}\setlength\itemsep{0pt}
\item A protocol that separates what an agent can discover from what it can hand on, and splits the handoff loss
into the change of agent, a single compression and repeated compression, using controls on identical situations
(\cref{sec:protocol}).
\item 56 admitted worlds in six themes with programmatic rules, fact lists and designer notes, and a
procedure that grades any note fact by fact (\cref{sec:worlds}).
\item Measurements on six models from two providers: a handoff loss for every model, an exchange rate of about two generations per life, and a loss that
follows each model's own frontier (\cref{sec:results}).
\item The cause, established by intervention: most of the loss is produced by summarising summaries, and is
removed by writing each summary from the record (\cref{sec:notes,sec:interventions}).
\end{itemize}

\section{Related Work}
\label{sec:related}

Our study sits where four lines of work meet: handoff between agents, context compaction, agents that learn from
their own experience, and transmission chains. \Cref{tab:related} compares it with the closest study in each on
five properties of the design.

\begin{table}[h]
\caption{\textbf{What this work measures that the closest prior studies do not.} \emph{Induced}: the knowledge
handed over must first be worked out by the agent, not supplied to it. \emph{Graded}: the handed-over text is
checked against a ground truth, statement by statement. \emph{Chain}: more than one handoff in succession.
\emph{Record}: a baseline that passes the history uncompressed. \emph{Once vs.\ repeated}: compressing
once is separated from compressing a compression.}
\label{tab:related}
\centering\small
\newcommand{\y}{\checkmark}\newcommand{\n}{$\times$}
\setlength{\tabcolsep}{4pt}
\resizebox{\linewidth}{!}{%
\begin{tabular}{@{}llccccc@{}}
\toprule
Study & What is passed on & Induced & Graded & Chain & Record & Once vs.\ rep. \\
\midrule
Human chains \citep{tessler2021growing} & how to play a new game & \y & \n & \y & \y & \n \\
LLM telephone game \citep{perez2024telephone} & a text to be retold & \n & \n & \y & \n & \n \\
Handoff Debt \citep{kc2026handoff} & an unfinished coding task & \n & \n & \n & \y & \n \\
Lost in Compaction \citep{wang2026lost} & a user's instructions & \n & \y & \n & \y & \n \\
Ground Truth First \citep{spencer2026groundtruth} & facts stated in a dialogue & \n & \y & \n & \n & \n \\
LongMemEval \citep{wu2024longmemeval} & facts stated in a dialogue & \n & \y & \n & \y & \n \\
ARC-AGI-3 \citep{arcagi3} & nothing (one agent) & \y & \n & \n & \n & \n \\
\midrule
\textbf{This work} & rules the agent worked out & \y & \y & \y & \y & \y \\
\bottomrule
\end{tabular}}
\end{table}

\paragraph{Handoff between agents.}
Multi-agent frameworks pass work between roles as text \citep{wu2023autogen,hong2023metagpt,li2023camel}, and an
analysis of their failures attributes a large share to what is lost or misread between agents
\citep{cemri2025mast}. The closest study to ours is \citet{kc2026handoff}, who interrupt coding agents on
SWE-bench tasks \citep{jimenez2023swebench} and let a successor continue from the repository alone, the raw
trace, free-form notes, or structured notes; see also \citet{santos2026esaa} and \citet{jarmak2026reliable}.
Handoff material cuts the successor's work, while effects on solve rate are smaller and depend on the model.
That study covers one handoff of a task whose specification is known. In this work, we study a chain of handoffs
of knowledge that the agents must first induce, which is what lets us compare a summary of the evidence with a
summary of a summary, and check each statement in a note against the truth.

\paragraph{Context compaction.}
Long contexts are used unevenly \citep{liu2023lost}, and agents that outgrow their window summarise or evict
history \citep{packer2023memgpt,kang2025acon,cim2026parallel,semenov2026eviction,li2026compactionrl}. A common
design keeps a running summary and folds new turns into it, so that each summary is written from the last one
\citep{packer2023memgpt}. Recent evaluations ask what a summary drops: instructions meant to govern the rest of a
session \citep{wang2026lost}, state that the agent then spends extra calls to recover
\citep{liu2026compressioncost}, or commitments and facts that should have been kept
\citep{kwon2026reclaim,trukhina2026commitments,wang2026forget}. In these studies the content at risk is
information the agent was given or steps it took, and it is compressed once. In this work, the content is what
the agent had to work out, and we separate compressing once from compressing repeatedly. We find most of the
loss in the second, which is the configuration a running summary produces.

\paragraph{Agents that learn from their own experience.}
Agents improve across episodes by writing reflections \citep{shinn2023reflexion}, extracting insights
\citep{zhao2023expel}, storing skills or workflows \citep{wang2023voyager,wang2024awm}, or maintaining a
structured memory \citep{park2023generative,xu2025amem,chhikara2025mem0}; surveys and benchmarks of such memory
are now numerous \citep{du2026memsurvey,maharana2024locomo,wu2024longmemeval,hu2025memoryagentbench,
ai2025memorybench,li2026ememb,zheng2025lifelong}. What gets written is not always right: a wrong record is
copied into later tasks \citep{xiong2026memory}, a stored conclusion shapes the observations that then confirm
it \citep{liao2026belief}, agents rely on stored content that current evidence contradicts
\citep{hu2026trustgap,chao2026stale,singh2026belief}, and early mistakes cascade \citep{zhu2025fail}.
\citet{spencer2026groundtruth} audit what a memory writer stores against a script fixed in advance and find that
write quality predicts downstream accuracy. We share the emphasis on the write step and on ground truth that
exists before any text does. In this work, we hold the experience fixed and vary only what the writer can see
and what the reader receives, so that the cost of writing is separated from the cost of not knowing.

\paragraph{Transmission chains and iterated learning.}
Passing material along a chain of people is a classic method for studying what communication preserves
\citep{bartlett1932remembering,mesoudi2008multiple}. The accumulation of improvements across generations, the
``ratchet'' \citep{tomasello1999cultural}, has been reproduced in the laboratory
\citep{kirby2008cumulative,caldwell2008experimental}, and models of iterated learning predict that when each
learner sees only its predecessor's output, what the chain retains drifts toward what the learners already
expect and away from the evidence \citep{griffiths2007language,beppu2009iterated}. Our protocol follows
\citet{tessler2021growing}, in which chains of eight people played unfamiliar video games with two lives each
and left a message for the next player; the chains improved along a trajectory close to that of individuals
with unlimited lives. Chains of language models have been used to study the drift of text toward attractors
\citep{perez2024telephone,perez2024populations}, the evolution of cooperation \citep{vallinder2024cultural},
the transmission of invented languages \citep{stengeleskin2026glossogen} and the degradation of models trained
on their own output \citep{guo2026collapse}; \citet{zhuang2025ratchet} and \citet{lu2026utt} study how knowledge
written by one model can raise another. None of these measures the transmission of task knowledge against known
rules. In this work, we do, and we add two controls that a study with people cannot easily run, which separate
what the chain loses from what the message loses.

\paragraph{Interactive benchmarks with hidden rules.}
A growing set of benchmarks tests whether a single agent can work out how an unfamiliar environment behaves
\citep{chollet2019measure,arcagi3,warrier2025autumn,battleday2026dig,zheng2025newton,chen2026causalgame},
alongside game suites that test play more broadly
\citep{liu2023agentbench,costarelli2024gamebench,hu2025lmgame,hua2026ptcg}. That ability corresponds to one
segment of what we measure, the gap between an agent that keeps its whole history and an agent that is told the
rules. In this work, the worlds are built for the other segment: they are small enough that the rules fit in a note,
so that what remains to be explained is why the note does not contain them.

\section{Worlds}
\label{sec:worlds}

A world is a small interactive program with hidden rules. For the comparison in \cref{sec:protocol} to mean
anything, three things must hold. An agent that is told the rules must succeed, so that failure can be
attributed to missing knowledge. An agent that is told nothing must fail within one life, so that there is
something to learn. And the rules must form a finite list of checkable facts, so that any text about the world
can be graded.

\begin{figure}[t]
  \centering
  \includegraphics[width=\linewidth]{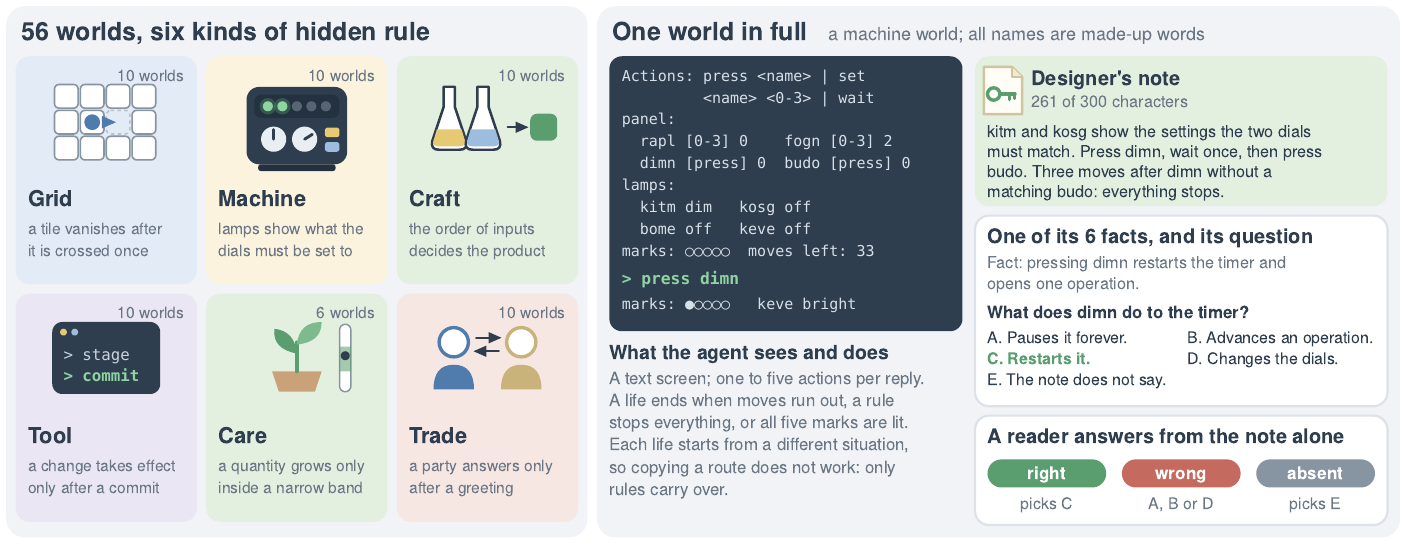}
  \caption{\textbf{The worlds.} \emph{Left}: six themes that differ in how a rule shows itself, with an example
  rule and the number of admitted worlds. \emph{Right}: one world in full. The agent reads a text screen and
  replies with actions; the designer's note states the rules within 300 characters; each of the world's facts
  has a multiple-choice question, and a reader that sees only a note answers it, which grades the note fact by
  fact as right, wrong or absent.}
  \label{fig:worlds}
\end{figure}

\paragraph{Task.}
A world is a deterministic program. In one \emph{life} the agent reads a text screen, replies with one to five
short text actions, and reads the screen that results, until the life ends (\cref{fig:worlds}, right). A life
ends when the agent runs out of moves, when it triggers a rule that stops everything, or when it reaches the
goal. The agent is never told the goal. A row of five \emph{marks} on the screen shows how far it has come, the
fifth mark is the goal, and the score of a life is the highest mark reached.

\paragraph{An example.}
In the world of \cref{fig:worlds} the screen shows two dials, two buttons and four lamps. Two lamps display, as
off, dim, bright or blinking, the settings that the dials must be given. One button opens a timed operation and
earns the first mark; the other completes the operation, but only if the dials match and one further move
has been made first. Three moves after the first button without a correct completion, everything stops. Four completions
light all marks. None of this is on the screen. An agent that explores sees lamps change and lives end, and has
to work out which of its actions were responsible.

\paragraph{Instances.}
Each world has sixteen \emph{instances} that share its rules and differ in layout, starting values and the
sequence of actions that reaches the goal. Replaying the solution of one instance on another reaches the goal
in at most 5\% of cases. Consecutive lives always use different instances, so a route is worth nothing to the
next life and a rule is worth everything: a note is useful only if it states something that holds in every
instance.

\paragraph{Themes.}
Worlds come in six themes that differ in how a rule shows itself (\cref{fig:worlds}, left). In a grid world most
rules are visible the moment they fire. In a machine world the meaning of an indicator has to be inferred from
several configurations, and one life shows only a few of them. A world combines three to five mechanisms from
its theme's library, and its names are pseudo-words drawn at random, with a list of meaning-bearing words banned
from screens, so that nothing can be recalled from pretraining and nothing is given away by a name.

\paragraph{What a world ships with.}
Every world comes with three things besides its program. A \emph{solver} plays any instance to the goal from the
rules. A list of \emph{facts} states the rules one at a time, each with a multiple-choice question whose options
include ``the note does not say''; the 56 worlds have 323 facts, five to eight per world,
about what the goal is, what ends a life, how a mechanism works, in what order things must be done, and in what
quantity. And a \emph{designer's note} states the rules in at most 300 characters (median 253).
\Cref{tab:stats} gives the statistics.

\begin{table}[t]
\caption{\textbf{Statistics of the 56 admitted worlds.} Ranges are minimum to maximum, with the median in
brackets.}
\label{tab:stats}
\centering\small
\resizebox{\linewidth}{!}{%
\begin{tabular}{@{}lr@{\qquad}lr@{}}
\toprule
Worlds (grid, machine, craft, tool, care, trade) & 10, 10, 10, 10, 6, 10 & Facts in all & 323 \\
Instances per world & 16 & Facts per world & 5--8 [6] \\
Moves in a life & 13--60 [33] & \quad about a mechanism / an order & 78 / 69 \\
Characters on the opening screen & 97--298 [187] & \quad about the goal / a danger & 65 / 64 \\
Mechanisms per world & 3--5 [4] & \quad about a quantity / an exception & 36 / 11 \\
Marks & 5 & Designer's note, characters & 194--300 [253] \\
\bottomrule
\end{tabular}}
\end{table}

\paragraph{Construction and admission.}
Worlds were built in three stages. \emph{Stage I.} For each theme we wrote an engine, a library of mechanisms
that can be combined, together with a written specification of what a world must satisfy. \emph{Stage II.}
Sixty worlds, ten per theme, were drafted against the specification with the help of a coding agent, each with
its solver, facts and designer's note. \emph{Stage III.} A world is admitted only if it passes automatic checks
and a model check. The automatic checks require that the solver reaches the goal on every instance using
55--70\% of the available moves, so that an informed agent has slack and an uninformed one cannot try
everything; that 2{,}000 lives of uniformly random actions reach the goal at most twice and the fourth mark at
most 1\% of the time; that at least 10\% of those lives are ended by a rule, and that no first move ends a life;
that screens are deterministic; and that the facts, questions and note are well formed. The model check uses
one reference model (deepseek-v4.1-flash): with the designer's note it must reach the goal in at least 8 of 10
lives, and without a note in at most 2 of 10. Of the 60 worlds, 56 were admitted; four care
worlds could not be brought to pass and were dropped. On the admitted worlds the reference model reaches the
goal in 0.6\% of lives without a note and 95.8\% with the designer's note.

\paragraph{Why these properties matter.}
Three consequences of the design carry the rest of the paper. \emph{What should be handed over is known.} The
facts exist before any agent plays, so a note can be graded against them, and what a writer left out can be
told apart from what it got wrong. \emph{The rules fit in a note.} A designer writes them in the 300 characters
that our agents are allowed, so when an agent's note does not contain them, the limit is not the reason.
\emph{Acting on the rules is within reach.} With the designer's note the reference model, the weakest we test,
reaches the goal in almost every life, so a low score reflects what an agent knows, not how well it plays.
Writing such a note is itself not trivial: several designer notes that stated every rule failed the model check
until they were reworded, a first sign that 300 characters another agent can act on are hard to write.

\section{Protocol}
\label{sec:protocol}

\begin{figure}[t]
  \centering
  \includegraphics[width=\linewidth]{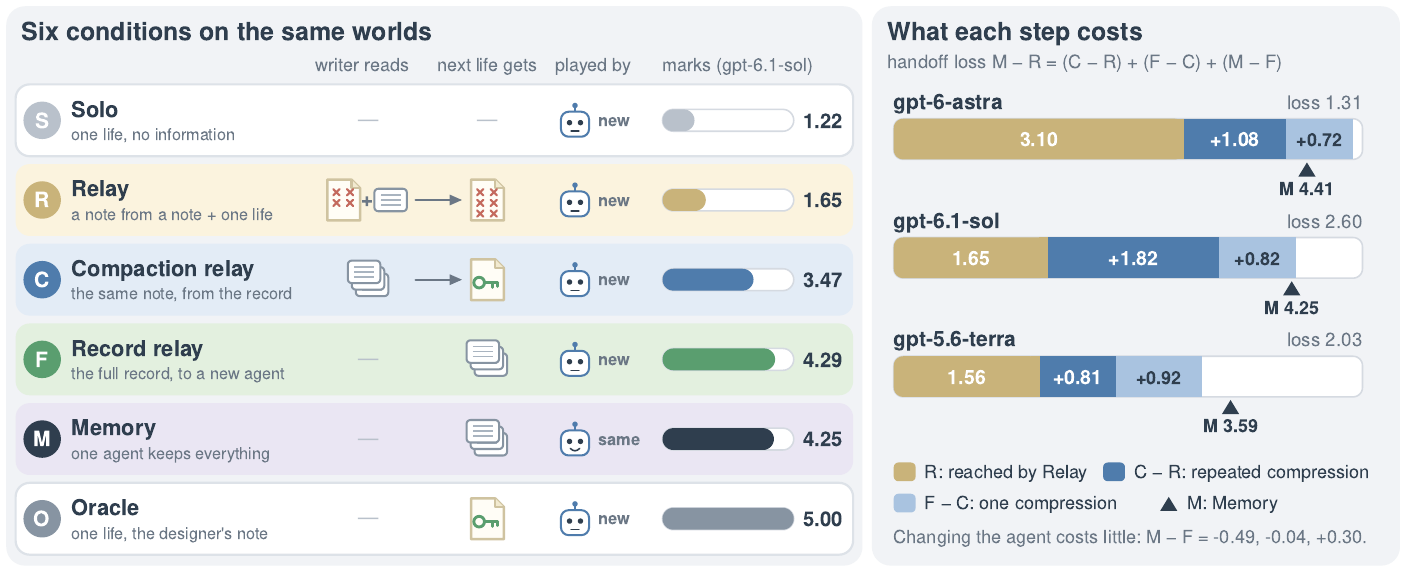}
  \caption{\textbf{The protocol and the decomposition it yields.} \emph{Left}: the six conditions, with what the
  writer of a note may read, what the next life receives, and whether the agent changes; bars give the score of
  gpt-6.1-sol (Solo and Oracle: all lives; the four sequences: lives 5--8). \emph{Right}: the handoff loss split
  into its parts for three models, each on the ten worlds where its loss is largest. Letting the writer read the
  full record regains most of the loss for the two stronger models; passing the record uncut regains the rest.}
  \label{fig:conditions}
\end{figure}

The protocol asks one question of a model on a world: how much of what an agent learns over several lives
reaches the lives that follow, and where the rest is lost. We answer it with six conditions arranged so that
neighbouring conditions differ in exactly one respect (\cref{fig:conditions}); each difference in score then
has a single cause.

\paragraph{Setting.}
A \emph{sequence} is eight lives of one world, played one after another on different instances. Life $g$ leaves
a \emph{record} $h_g$, the screens it showed and the actions taken in it; the agent's reasoning is not part of
the record. A \emph{note} $n_g$ is a text of at most $L=300$ characters, written after life $g$ in reply to a fixed
request: ``Write a note of at most 300 characters for the next player, who will meet the same world in a
different starting situation.'' Longer notes are cut at the limit. A condition is defined by two things: what
the agent that plays life $g$ receives from the past, and what the writer of a note has in front of it.

\paragraph{Six conditions.}
Two conditions are single lives that bound the scale. In \textbf{Solo} the agent receives nothing, which gives
the score of an agent that has learned nothing. In \textbf{Oracle} it receives the designer's note, which gives
the score of an agent that has been told everything. The other four are sequences.

\textbf{Relay.} Eight agents play one life each. The agent of life $g$ receives $n_{g-1}$ and nothing else, and
writes $n_g$ from what it has seen: $n_{g-1}$ and its own life $h_g$. From the second life on, every note is
therefore written from a note. This is the handoff under study.

\textbf{Compaction relay.} As Relay, with one change: when the agent of life $g$ writes its note, it is also
shown the records $h_1,\dots,h_{g-1}$ of all earlier lives, which it did not see while playing. The next agent
still receives 300 characters. The limit, the request and the reader are those of Relay; only the writer's view
differs, so the comparison isolates the effect of writing a summary from a summary instead of from the evidence.

\textbf{Record relay.} Eight agents play one life each, and the agent of life $g$ receives $h_1,\dots,h_{g-1}$
in place of a note. No note is written and nothing is compressed. Against Compaction relay this isolates the
cost of the 300 characters themselves.

\textbf{Memory.} One agent plays all eight lives in a single conversation and keeps everything it has seen and
said. This is the unlimited-lives control of \citet{tessler2021growing}. Against Record relay it isolates the
cost of changing the agent: both have every screen and action, and only one of them lived through them.

To hold everything else fixed, life $g$ of sequence $k$ uses the same instance in all four sequence conditions,
so that they meet the same situations in the same order, and screens and replies have the same format in every
condition.

\paragraph{Measures.}
The score of a life is the highest mark reached, from 0 to 5. For the sequences we average the score over lives
5--8, after learning has had time to occur, and show the earlier lives as learning curves. Let $S$, $R$, $C$,
$F$, $M$ and $O$ be the mean scores under Solo, Relay, Compaction relay, Record relay, Memory and Oracle. The
\emph{handoff loss} is $M-R$, what an agent with its whole history has that a chain of notes does not. Because
neighbouring conditions differ in one respect, it splits exactly into three parts:
\begin{equation*}
\underbrace{M-R}_{\text{handoff loss}} \;=\;
\underbrace{(M-F)}_{\text{change of agent}} \;+\;
\underbrace{(F-C)}_{\text{one compression}} \;+\;
\underbrace{(C-R)}_{\text{repeated compression}} .
\end{equation*}
Two further gaps place the loss on the scale: $R-S$, the \emph{note value}, is what a chain gains over an
uninformed life, and $O-M$, the \emph{discovery gap}, is what an agent fails to work out even with its whole
history. Because the scale differs between models, we also report the loss in two relative forms: as a share
of what was learned, $(M-R)/(M-S)$, and as an exchange rate, the life at which the Memory curve first reaches
the score of the eighth Relay life. The unit of analysis is the world: scores are averaged within a world
first, and 95\% confidence intervals come from resampling worlds.

\paragraph{Grading notes.}
Scores show how much is lost; the notes show what. A fixed reader model receives a note and the fact questions
of its world, and nothing else, and selects an option for each fact, which is then \emph{right}, \emph{wrong} or
\emph{absent} (the reader chose ``the note does not say''). Since a sequence yields a note after every life,
this gives the fate of each fact: when it first appears, whether it is kept, and whether a wrong version is
ever corrected. Two reader models agreed on 91.1\% of 1{,}591 judgements (Cohen's
$\kappa=0.81$), and in a sample of their disagreements the default reader was correct in 10 of
10. A second
reader, given the true rules, splits each note into claims and labels each as a rule that holds in every game,
a false claim, or a record of a single game.

\paragraph{Variants of the note.}
To test whether the loss is a matter of the limit or of the request, we vary both in Relay. We set $L$ to 150,
600, 1{,}500 and 3{,}000 characters. And at fixed $L$ we change the request to the writer in three ways: to
start from the inherited note and revise it, as in \citet{tessler2021growing}; to say for each rule whether it
was verified and to record observations it could not explain; or to write a ledger of what is confirmed, what
has been ruled out, and what is still open.

\paragraph{Models and runs.}
We evaluate six models from two providers (\cref{tab:main}), with reasoning effort set to high where
the option exists and all other settings at their defaults. Solo, Oracle, Relay and Memory cover all
56 worlds: 5 lives per world for the first two and 3 sequences per world for the last two
(5 Relay chains for deepseek-v4.1-flash). The other conditions cost more and were run where they
are informative. Compaction relay covers all 56 worlds for one model, 29 for a second and
10 for a third; Record relay and the note variants cover, for each of three models, the 10
worlds where its handoff loss is largest. In all, the experiments comprise more than 25{,}000 lives.
A request that returns no response is retried, since the model has seen nothing; a life whose requests keep
failing is recorded as missing and is not scored as zero.

\section{Results}
\label{sec:results}

We report four findings in turn: every model loses at a handoff (\cref{sec:curve}); most of what it loses is
lost to compressing repeatedly, not to the limit and not to the change of agent (\cref{sec:decomp}); the loss
falls on the worlds a model can only just learn (\cref{sec:where}); and the notes fall short because the
writer has not yet arrived at the rules (\cref{sec:notes}).

\subsection{Every model loses at a handoff}
\label{sec:curve}

\begin{table}[t]
\caption{\textbf{Scores under the four main conditions and the handoff loss}, 56 worlds. Relay and Memory are
means over lives 5--8. \emph{Share}: the loss as a fraction of what Memory gains over Solo. \emph{Rate}: the
Memory life whose score equals that of the eighth Relay life. \emph{Goal}: lives 5--8 that reach the fifth
mark. Brackets are 95\% intervals from resampling worlds. $^\dagger$31 worlds.}
\label{tab:main}
\centering\small\setlength{\tabcolsep}{3.6pt}
\begin{tabular}{@{}l cccc @{\hspace{9pt}} l@{\hspace{3pt}}l c c @{\hspace{9pt}} cc@{}}
\toprule
 & \multicolumn{4}{c}{Mean marks (0--5)} & \multicolumn{4}{c}{Handoff loss} & \multicolumn{2}{c}{Goal (\%)} \\
\cmidrule(r{9pt}){2-5}\cmidrule(r{9pt}){6-9}\cmidrule{10-11}
Model & Solo & Relay & Mem. & Oracle & \multicolumn{2}{l}{$M-R$} & Share & Rate & Relay & Mem. \\
\midrule
gpt-6-astra   & 2.21 & 4.12 & 4.57 & 4.99 & \lossbar{0.44} & \textbf{0.44} {\scriptsize[0.24, 0.66]} & 19\% & 4.0 & 65 & 82 \\
gpt-6.1-sol   & 1.96 & 3.77 & 4.41 & 4.96 & \lossbar{0.64} & \textbf{0.64} {\scriptsize[0.31, 0.96]} & 26\% & 4.1 & 57 & 74 \\
gpt-5.6-terra & 1.41 & 2.40 & 3.07 & 4.68 & \lossbar{0.67} & \textbf{0.67} {\scriptsize[0.45, 0.89]} & 40\% & 3.9 & 17 & 27 \\
gpt-5.6-luna  & 1.30 & 1.94 & 2.52 & 4.61 & \lossbar{0.59} & \textbf{0.59} {\scriptsize[0.35, 0.84]} & 48\% & 2.5 & 7 & 15 \\
gpt-6-luna    & 1.25 & 1.87 & 2.13 & 4.62 & \lossbar{0.26} & \textbf{0.26} {\scriptsize[0.07, 0.46]} & 30\% & 3.9 & 4 & 10 \\
deepseek-v4.1-flash$^\dagger$ & 1.12 & 1.79 & 2.22 & 4.89 & \lossbar{0.43} & \textbf{0.43} {\scriptsize[0.20, 0.65]} & 39\% & 3.0 & 6 & 14 \\
\bottomrule
\end{tabular}
\end{table}

\begin{figure}[t]
  \centering
  \includegraphics[width=\linewidth]{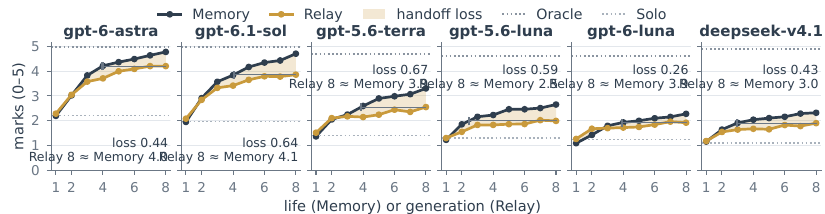}
  \caption{\textbf{Learning curves of six models} (56 worlds; deepseek-v4.1-flash: 31). Memory and Relay meet the
  same instances in the same order. The shaded band is the handoff loss. The grey tick marks the Memory life
  whose score the eighth Relay life reaches: a chain of eight agents stands where one agent stood after about
  four lives.}
  \label{fig:learning}
\end{figure}

\textbf{The conditions order the same way for every model: $S<R<M<O$} (\cref{tab:main}). With the designer's
note every model scores at least 4.61 of 5, so all of them can act on the rules. With no information none
exceeds 2.21. Models differ mostly in the discovery gap $O-M$: 0.42 and 0.55 marks for the two strongest
against 1.61 to 2.67 for the other four, which do not work out most worlds in eight lives.

\textbf{The handoff loss is positive for all six models.} Every 95\% interval excludes zero. For gpt-6.1-sol
the late lives of a chain reach the goal 57\% of the time and those of the memory agent 74\%. For the four
weaker models the goal is reached too rarely for a rate to show much (4--17\% against 10--27\%), and the loss is
visible in marks: a benchmark that scored completion alone would have missed it.

\textbf{Eight generations are worth about four lives.} Reading the score of the eighth Relay life off the
model's own Memory curve (\cref{fig:learning}), a chain of eight stands where a single agent stood after 4.0,
4.1, 3.9 and 3.9 lives for gpt-6-astra, gpt-6.1-sol, gpt-5.6-terra and gpt-6-luna, and after 2.5 and 3.0 for
gpt-5.6-luna and deepseek-v4.1-flash. Models whose absolute scores differ by a factor of two pay a similar
rate. The Relay curve keeps rising to the last life, so on average a chain is a slower learner, not a stalled
one.

\textbf{A fifth to a half of what is learned does not pass.} As a share of what Memory gains over Solo, the
loss is 19\% and 26\% for the two strongest models and 30--48\% for the others. The ranking by share is not
the ranking by score: gpt-6-luna learns least and loses 30\%, gpt-5.6-luna learns more and loses 48\%.
Learning from experience and handing it over are separable abilities.

\subsection{Most of the loss comes from compressing repeatedly}
\label{sec:decomp}\label{sec:interventions}

\begin{figure}[t]
  \centering
  \includegraphics[width=\linewidth]{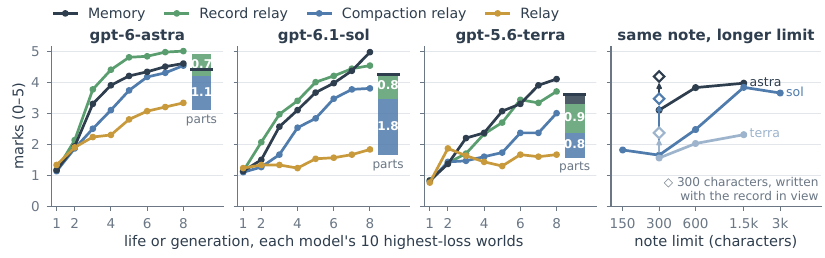}
  \caption{\textbf{Where the loss arises.} \emph{Left three panels}: the four sequence conditions on each
  model's ten highest-loss worlds. The bar splits the distance from Relay up to Memory (black line) into what
  is regained when the writer reads the full record (blue, repeated compression) and when the record is passed
  on uncut (green, one compression); what is left is the change of agent. \emph{Right}: Relay with a longer
  note. Diamonds: Compaction relay at 300 characters, which does about as well as a Relay note five times as
  long.}
  \label{fig:decomp}
\end{figure}

\begin{table}[t]
\caption{\textbf{The handoff loss split into its three parts.} Top: each model's ten highest-loss worlds, all
four sequence conditions. Bottom: every world on which Compaction relay was run. Brackets are 95\% intervals
from resampling worlds.}
\label{tab:decomp}
\centering\small\setlength{\tabcolsep}{4pt}
\begin{tabular}{@{}l r cccc @{\hspace{10pt}} lll@{}}
\toprule
 & & \multicolumn{4}{c}{Mean marks, lives 5--8} & \multicolumn{3}{c}{Parts of the loss $M-R$} \\
\cmidrule(r{10pt}){3-6}\cmidrule{7-9}
Model & Worlds & $R$ & $C$ & $F$ & $M$ & Repeated $C-R$ & Once $F-C$ & Agent $M-F$ \\
\midrule
gpt-6-astra   & 10 & 3.10 & 4.18 & 4.90 & 4.41 & \textbf{1.08} {\scriptsize[0.40, 1.84]} & 0.72 {\scriptsize[0.22, 1.29]} & $-$0.49 {\scriptsize[$-$1.00, $-$0.01]} \\
gpt-6.1-sol   & 10 & 1.65 & 3.47 & 4.29 & 4.24 & \textbf{1.82} {\scriptsize[1.16, 2.42]} & 0.83 {\scriptsize[0.28, 1.57]} & $-$0.05 {\scriptsize[$-$0.57, 0.53]} \\
gpt-5.6-terra & 10 & 1.56 & 2.37 & 3.29 & 3.59 & 0.81 {\scriptsize[0.23, 1.38]} & \textbf{0.92} {\scriptsize[0.14, 1.68]} & \phantom{$-$}0.30 {\scriptsize[$-$0.25, 0.84]} \\
\midrule
gpt-6.1-sol   & 56 & 3.77 & 4.16 & --- & 4.41 & \textbf{0.39} {\scriptsize[0.15, 0.64]} & \multicolumn{2}{l}{0.25 {\scriptsize[0.02, 0.49]} \;(the two together)} \\
gpt-6-astra   & 29 & 4.22 & 4.66 & --- & 4.76 & \textbf{0.45} {\scriptsize[0.14, 0.80]} & \multicolumn{2}{l}{0.09 {\scriptsize[$-$0.10, 0.33]} \;(the two together)} \\
\bottomrule
\end{tabular}
\end{table}

A handoff changes three things at once: the agent, the amount of text that survives, and how many times that
text has been rewritten. The three further conditions change them one at a time (\cref{fig:decomp},
\cref{tab:decomp}).

\textbf{Changing the agent costs nothing.} A fresh agent that is given the complete record of earlier lives
does as well as the agent that lived them: on the ten highest-loss worlds of each model, $M-F$ is $-0.05$ for
gpt-6.1-sol and $0.30$ for gpt-5.6-terra, with intervals that include zero, and $-0.49$ for gpt-6-astra, whose
fresh agents do slightly \emph{better}. Whatever is lost at a handoff, it is not lost because experience
cannot be read by another agent.

\textbf{Compressing once costs something.} When the writer has the full record in front of it and must fit
what it knows into 300 characters, the next life scores 0.72 to 0.92 marks below a life that receives the
record itself.

\textbf{Compressing repeatedly costs the most.} The same 300 characters, written by an agent that has only a
note and its own life, cost a further 1.82 marks for gpt-6.1-sol and 1.08 for gpt-6-astra: 70\% and 82\% of
their loss on these worlds. For gpt-5.6-terra this part is 0.81 marks, 40\% of its loss, and one compression
costs as much; the weaker model is also the weaker summariser of a full record. The result holds beyond these
worlds. On all 56 worlds, Compaction relay raises gpt-6.1-sol from 3.77 to 4.16, regaining 0.39 of its 0.64
marks of loss (interval 0.15 to 0.64); on the 29 worlds run for gpt-6-astra it regains 0.45 of 0.54.

\textbf{The limit is not what binds.} If 300 characters were too few for what must be said, a writer with the
full record would be no better off. It is: at the same limit it comes close to or beats an ordinary note of 1{,}500
characters (\cref{fig:decomp}, right; 3.47 against 3.83 for gpt-6.1-sol, 4.18 against 3.97 for gpt-6-astra,
2.37 against 2.31 for gpt-5.6-terra), and giving it 1{,}500 characters adds nothing (3.47). An ordinary note
does improve with room, from 1.65 at 300 characters to 3.83 at 1{,}500 for gpt-6.1-sol, with no further gain
at 3{,}000 (3.65). We read the longer note as a partial stand-in for the record: it has room to carry
observations along with conclusions.

\textbf{Asking for a different note does not help.} On the same worlds, gpt-6.1-sol scores 1.98 when asked to
revise the inherited note instead of writing a new one, as the human chains of \citet{tessler2021growing} did;
1.57 at 300 characters and 1.95 at 600 when asked for a ledger of what is confirmed, ruled out and open
(1.65 and 2.48 with the plain request); and 3.49 at 1{,}500 characters when asked to mark what was verified
and to record what it could not explain (3.83 with the plain request). No instruction to the writer recovers
what access to the record recovers.

\subsection{The loss falls where a model can only just learn}
\label{sec:where}

\begin{figure}[t]
  \centering
  \includegraphics[width=\linewidth]{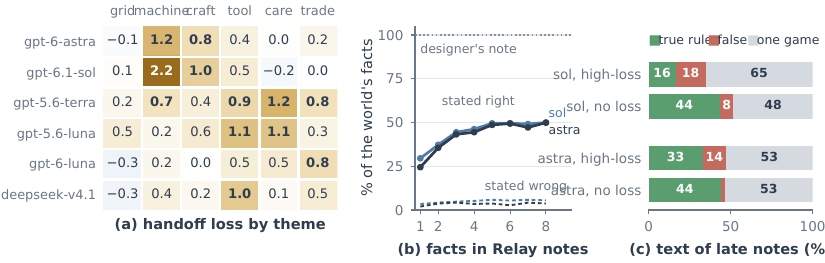}
  \caption{\textbf{Where the loss falls and what the notes contain.} (a) Handoff loss by theme. The two
  strongest models lose in machine and craft worlds and nowhere else; the others lose in the themes the
  strongest find easy. (b) Share of a world's facts that the Relay note states right (solid) or wrong
  (dashed) after each generation, against a designer's note that states all of them in the same space.
  (c) Text of the notes of lives 5--8 by what it says, on ten high-loss worlds and ten worlds without loss.}
  \label{fig:theme}\label{fig:fate}
\end{figure}

\textbf{For the two strongest models the loss sits in two themes.} gpt-6.1-sol loses 2.23 marks in machine
worlds and 0.96 in craft worlds, and gpt-6-astra 1.20 and 0.79; in grid, care and trade worlds neither loses
more than 0.22 (\cref{fig:theme}a). World by world the two models' losses have a rank correlation of 0.60, so
the loss is a property of the world and not an accident of one model's runs.

\textbf{A world learned at once loses nothing.} Grouping worlds by the life in which the memory agent first
reaches the goal, the 13 worlds that gpt-6.1-sol solves within two lives lose 0.03 marks, the 20 it solves in
lives 3--4 lose 0.62, and the 23 it solves later or never lose 1.01 (0.09, 0.65 and 0.56 for gpt-6-astra).
When one life is enough to work a world out, the note has only to state the result.

\textbf{The affected worlds move with the model.} A world that a model cannot learn even with memory has
nothing to hand on. The four weaker models score 1.2 to 1.9 with memory in machine worlds and lose only 0.2
to 0.7 there; their loss is largest in tool, care and trade worlds (up to 1.22), where the strongest models
lose little. Their world-by-world losses are uncorrelated with those of the two strongest models (rank
correlations $-0.21$ to $0.08$) and correlated with one another (up to 0.62). A stronger model does not
remove the loss. It moves it to the worlds that have just come within its reach.

\subsection{What the notes contain}
\label{sec:notes}

\textbf{Notes stop well short of the rules.} The share of facts that a Relay note states correctly rises from
30\% after the first life to 50\% by the fifth and stays there for gpt-6.1-sol, and from 24\% to 50\% for
gpt-6-astra (\cref{fig:fate}b). The designer's note states all of them within the same limit. In machine
worlds the late notes of gpt-6.1-sol state 17\%.

\textbf{What is written is kept; what is missing is rarely supplied.} A fact that is right in one note is
right in the next in 89\% of cases for gpt-6.1-sol and 87\% for gpt-6-astra. A fact that is absent appears in
the next note in 14\% and 15\%. A fact stated wrongly is corrected in 16\% and 20\% and repeated in 60\% and
49\%, but wrong statements stay rare (6\% and 4\% of facts in late notes). A chain is therefore good at
carrying a conclusion and poor at reaching one: each writer adds only what one life can show it.
Giving the writer the record changes exactly this. On the worlds graded so far (12 for gpt-6.1-sol, 9 for
gpt-6-astra, including the ten with the highest loss), late Compaction-relay notes state 47\% and 54\% of the
facts correctly, against 24\% and 40\% for Relay notes on the same worlds, and an absent fact is supplied in
the next note twice as often for gpt-6.1-sol (16\% against 8\%; 17\% against 12\% for gpt-6-astra).

\textbf{Where a chain fails, its notes hold fewer true rules, not more narration.} A second reader labels
each claim of a late note as a rule that holds in every game, a false claim, or a record of one game
(\cref{fig:fate}c). On the ten high-loss worlds the notes of gpt-6.1-sol are 16\% true rules, 18\% false
claims and 65\% single-game records; on ten worlds without loss they are 44\%, 8\% and 48\%. For gpt-6-astra
the single-game share is 53\% in both sets. The loss also does not depend on the kind of fact: goals, dangers,
mechanisms, orderings and quantities are all stated rarely in machine worlds (4--27\%) and all more often
elsewhere (37--69\%). Nothing crowds the rules out of the note, and no kind of rule is especially hard to
write down. The writer has not got as far as the rule.

\section{Discussion}
\label{sec:discussion}

We set out to measure what a handoff loses when the knowledge at stake has a known answer. The measurements
say that the loss is real for every model, that it follows each model's own frontier, and that most of it is
created by writing each note from the note before it. We discuss what explains this and what follows for
those who build and evaluate agents.

\paragraph{What explains the loss?}
We offer the following account. A note carries conclusions. A rule that needs several lives to work out is, for
most of a chain, not yet a conclusion: it exists as observations that no single agent could interpret. Those
observations are not in the note, because the writer did not know they mattered, and after the handoff they
are nowhere; the next agent cannot set them beside its own. This fits each of the main findings. The loss
vanishes in worlds learned within one life, absent facts are rarely supplied, a longer note helps while a
differently worded request does not, and the same writer does far better once it can read the record. It is
also what the theory of iterated learning leads one to expect of a chain in which each learner sees only the
previous learner's output \citep{griffiths2007language,beppu2009iterated}. In simpler terms, an agent cannot
hand over what it has not yet understood.

\paragraph{Implications for builders of agent systems.}
Systems that compact a context when it fills, keep a rolling summary, or brief a sub-agent from an earlier
brief all write each summary from the last one. In our measurements this is the expensive choice. At an
unchanged budget of 300 characters, writing each note from the full record regains 70\% and 82\% of the loss
for the two strongest models on the worlds where it is largest. We therefore recommend keeping the raw record
outside the working context and giving it to whoever writes the next summary: this costs storage, not context.
Where the record cannot be kept, a longer summary is the next best choice, and a more elaborate instruction to
the writer is not a substitute for either.

\paragraph{Implications for evaluation.}
Two properties of our measurements bear on how handoff should be evaluated. First, learning and handing over
are separable: the model that learns most from its own history is not the one that passes on the largest
share. A benchmark that scores only what one agent achieves will not see the difference. Second, completion
rates understate the loss. For models that seldom reach the goal under either condition, the loss is clear in
marks and invisible in success rate, in line with reports that compression leaves task completion unchanged
while raising the work needed to recover state \citep{liu2026compressioncost,kc2026handoff}. An evaluation of
handoff needs a graded measure of progress and a reference condition in which nothing is handed over.

\paragraph{Relation to human chains.}
\citet{tessler2021growing} found that chains of people, each leaving a message after two lives of a video
game, did about as well as individuals with unlimited lives. Our model chains do not. The settings differ in
ways that may matter: our successors meet a new instance, which removes the option of passing on a route, and
many of our worlds need more than one life to work out a rule. Whether people lose less than models on these
worlds is a question the protocol can answer directly, and the worlds are playable by people through a
graphical interface that renders the same screens.

\section{Limitations}
\label{sec:limitations}

A first limitation is that the worlds are synthetic and small. This is a deliberate choice: it is what makes
the rules a checkable list and lets a note be graded against them, which no study of handoff on real tasks
has been able to do. We do not claim that the size of the loss carries over to software engineering or other
real work. We claim that its parts can be measured and that their order is the same for the models we tested.

A second limitation is coverage. The four main conditions cover all worlds and six models; the two further
chains, which cost several times more, were run where the loss is concentrated, with Compaction relay
extended to all worlds for one model. The models come from two providers, and extending the comparison to
other families is a matter of compute.

A third limitation is the grading of notes, which relies on a reader model. Two readers agree on 91\% of
judgements, and none of our conclusions rests on the grades alone: each is also visible in the scores of the
agents that played from the notes.

\section{Conclusion}
We study what is lost when an agent's experience is replaced by its own short account of it, using worlds in
which what should be handed over is known. All six models we test lose at a handoff: a chain of eight agents
stands where one agent stood after about four lives. Controlled variants of the chain locate the loss. It
does not come from changing the agent, and it comes only in part from the limit on the note; most of it
arises because each note is written from the note before it. The same note, written by an agent that can read
the record, recovers most of what was lost. We hope the protocol and the worlds help in building agents that
can be trusted to brief their successors.

\section*{Code Availability}
The codebase used to construct the environments and run the experiments is not publicly available.

\bibliography{refs}
\bibliographystyle{iclr2027_conference}
\end{document}